# Self-enhancement of automatic tunnel accident detection (TAD) on CCTV by AI deep-learning

*Kyu-Beom Lee[1] and Hyu-Soung Shin[2]

[1] *Smart City & Construction Engineering (Geotechnical & Geo-Space Engineering), UST, Gyeonggi Province 10223, Korea*
[2] *Department of Future Technology and Convergence Research, KICT, Gyeonggi Province 10223, Korea*
[2] hyushin@kict.re.kr

## ABSTRACT

The deep-learning-based tunnel accident detection (TAD) system (Lee 2019) has installed a system capable of monitoring 9 CCTVs at XX site in November, 2018. The initial deep-learning training was started by studying 70,914 labeled images and label data. However, sunlight, the tail light of a vehicle, and the warning light of the working vehicle were recognized as a fire, and many pedestrians were detected in the lane of the tunnel or a black elongated black object. To solve these problems, as shown in Fig. 1, the false detection data detected in the field were trained with labeled data and reapplied in the field. As a result, false detection of pedestrians and fire could be significantly reduced.

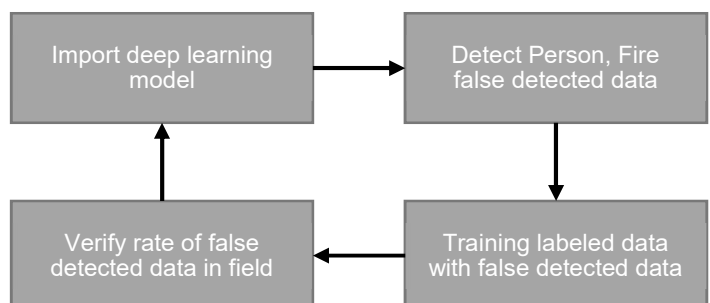

Fig. 1 The procedure of deep learning model training including false detected data

## 1. INTRODUCTION

An accident in a road tunnel can be defined as a fire, a person, a stoppage, or a car driving the wrong way (MOLIT 2016). If there was an accident in the road tunnel, it would be difficult for the driver to cope with the lack of evacuation space and limited visibility. Therefore, it should be possible to monitor the circumstance of road tunnels by installing CCTV in the tunnel, and the Ministry of Land, Infrastructure, and Transport of Korea recommends installation of a video accident detection system using CCTV. As a result, in November of 2018, the deep-learning-based video accident detection system was developed. Unlike the existing algorithm-based system, the deep-learning-based video accident detection system can adapt to the unclear and dark image environment of the tunnel itself. By learning the verified data set, the deep-learning-based video accident detection system can be employed in a tunnel site (Lee 2018).

---
[1] Graduate Student
[2] Professor

The deep-learning-based video accident detection system uses a deep-learning object detection network with the input of pre-processed still images as the first step. The deep-learning model classifies objects such as a car, a fire, or person and regresses the bounding box, represented by data in the form of a rectangular box.

The second step uses an object-tracking algorithm, simple online and real-time tracking (SORT), which is based on the car object detected in the deep-learning object detection network. The object-tracking process uses the car objects detected in the previous frame period and the current frame period. When there is a pair of objects whose overlapping area ratio is equal to or more than a predetermined value, the object number can be given and detected as the same object (Alex 2016).

The final steps, stoppage and wrong way driving, were performed for a pair of car objects having the same object ID in the previous frame period and the current frame period, and stoppage and wrong way driving were judged for a longer frame period than the object detection process. When the overlap area ratio was 0.9 or more, the stoppage was opposite to the tunnel direction, and when the overlapped line length ratio was less than 0.75, wrong way driving was judged.

The tunnel accident detection (TAD) system of this paper defines the deep-learning model obtained by a training data set with 70,914 still images and labeling data as model A. In November 2018, TAD based on model A started to operate at XX site.

However, since the TAD system was not a complete system, it was possible to detect not only the correct answers but also incorrect answers. Those types of results can be divided into 4 types as shown in Fig. 2.

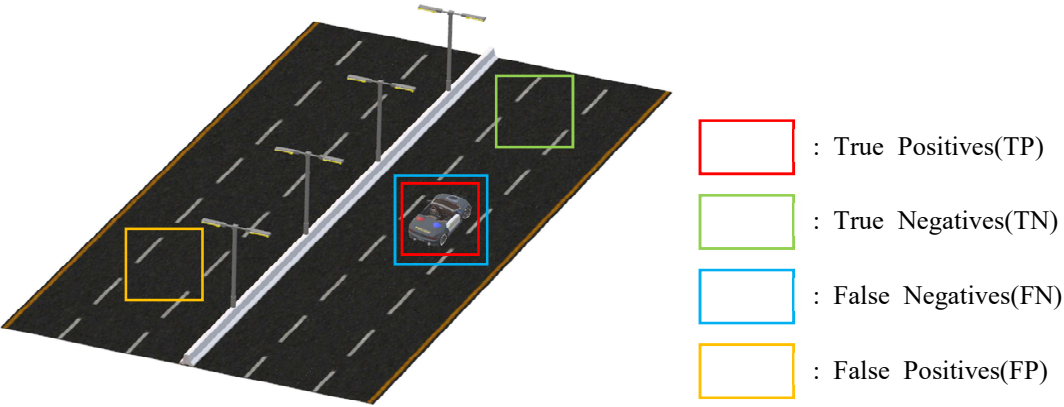

Fig. 2 4 types of prediction results for a car example

Fig. 2 is a graphical representation of the 4 types detected by the TAD system (Davis 2006). In Fig. 2, the target of detection of the car object is shown as an example. First, the bounding box of the red rectangle is defined as true positive (TP), and the car is correctly detected; TP is the primary target to detect for the TAD system. True negative (TN), represented by the green bounding box, refers to a case where an object is detected as an incorrect answer, not a car. The more TNs, the more reliable the TAD system. False negative (FN), represented by the blue bounding box in Fig. 2, or the case where it was not expressed at all, and when the correct answer is not detected. Finally, false positive (FP), represented by the yellow bounding box, is actually a wrong answer, but the system detects the correct answer.

The goal of the TAD is to maximally recognize TPs. However, when the performance of the TAD system was actually monitored, TP results were a minority, and there were many cases in which FPs were detected. This situation makes the TAD system unreliable for when operators would operate the TAD system.

In this paper, reducing the detection of FPs is the second objective of the TAD system. For this purpose, the aim is to reduce the FPs detected in fire and person objects except the cars that are detected most accurately in the deep-learning object detection network.

## 2.DEEP LEARNING SELF-ENHANCEMENT TRAINING METHOD

The deep-learning self-enhancement training method including FP data can be performed by first applying the TAD system to the field and then collecting the FP data detected in the field. In this work, the TAD system was applied to the field in November 2018 was applied. Then the FP data of the TAD system was identified, and the status of the FP data was monitored. In addition, the deep-learning self-enhancement training method including the FP data is presented, and the advantages of the deep-learning training method will be demonstrated in comparison with the deep-learning method using only the existing labeling data.

### 2.1. TAD system monitoring

The TAD system detects not only TPs but also TN, FP, and FN types. However, this type of identification cannot be confirmed by the TAD system itself; rather, it must be manually verified to classify and navigate the necessary types.

Since this work used FP data, the procedure of collecting FP data at a tunnel site is illustrated in Fig. 3.

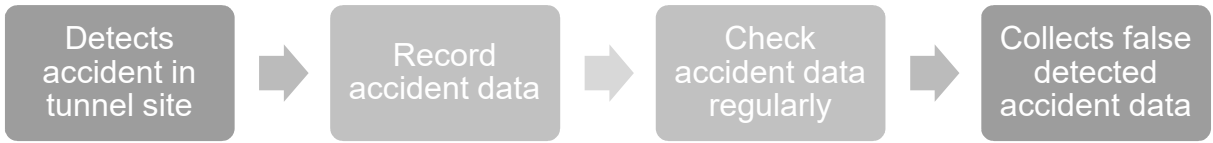

Fig. 3 The collection process for false detected accident data on TAD system

As seen in Fig. 3, the TAD system detects an accident by CCTV at a tunnel site, and then it records the accident data, which consists of a still image and bounding box data. Then, developers of TAD system regularly visit the site to identify the detected FP data up to the point of a visit and collect FP data from the recorded accident data.

In the object-detection stage of the TAD, car, person, and fire objects were detected. Car objects were detected more accurately than person or fire objects, but fire or person objects occasionally were detected, and most of them were FPs, not actual accidents.

Therefore, in this work fire and person objects were selected to detect FPs rather than cars, which were detected relatively accurately, and the numbers of false fires and false persons that were detected by 9-channel CCTV for 24 days after the first installation of TAD system were recorded.

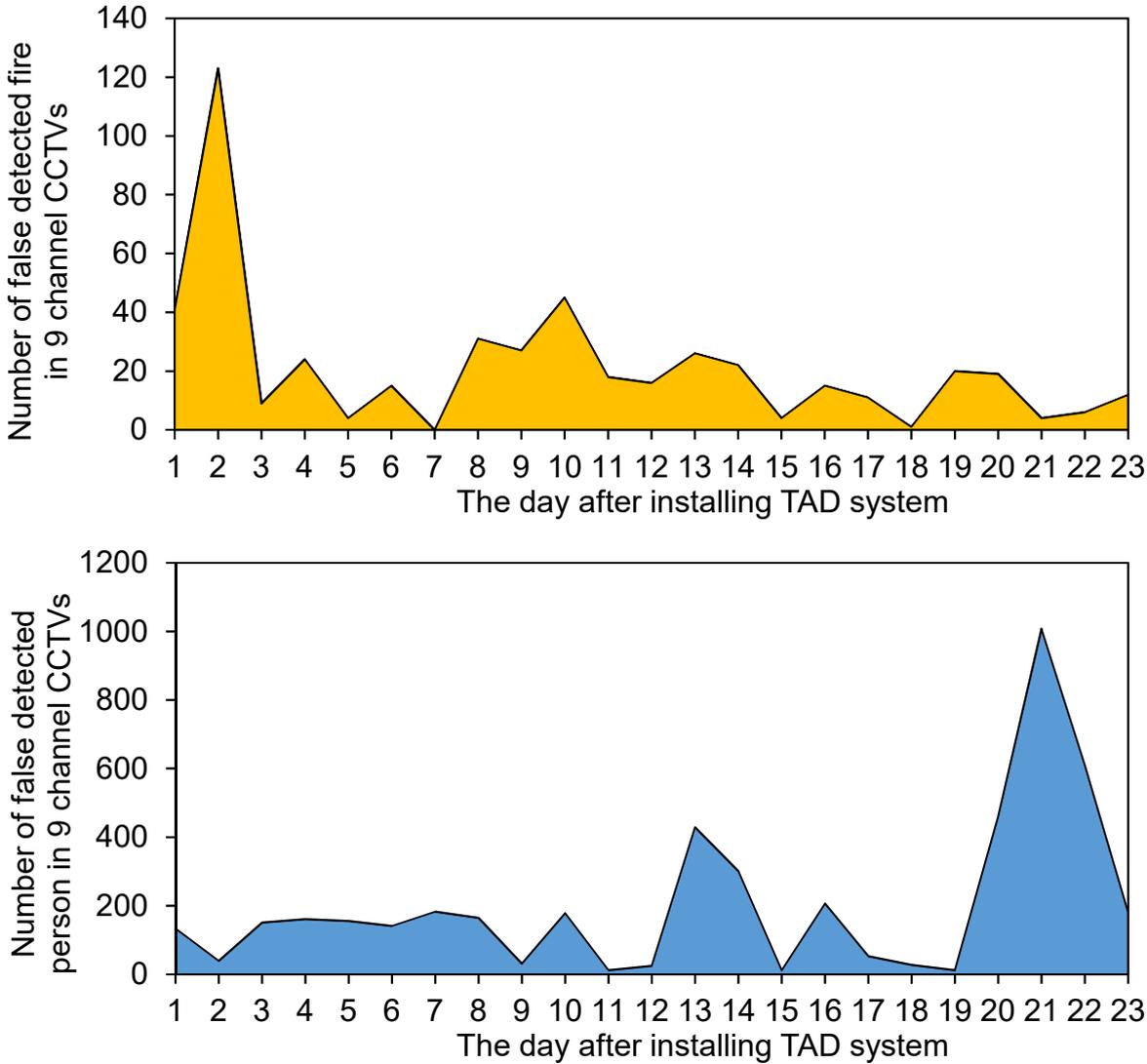

Fig. 4 The number of false detected fire, person on TAD system per day

Fig. 4 shows the number of accidents detected on 9 channels for fire and person objects after 9-channel CCTV surveillance began with the installation of the TAD system in November 2018. Fire objects were detected as FP in a maximum of 120 and usually 0 to 40 ranges after the TAD system was installed, and the frequency of fire alarms was 0 to 2 per hour. The number of falsely detected person objects was much greater than that of fire, with a maximum of 1000 person objects detected and an average of 0 to 400 person objects detected as FPs. Zero to 15 falsely detected person objects were detected per hour, and an alarm was displayed to the tunnel manager. As shown in Fig. 4, the TAD system based on the deep-learning model, which is trained only by the labeling data, caused a large number of false alarms for person and fire objects because the labeling data is similar to false-detected data.

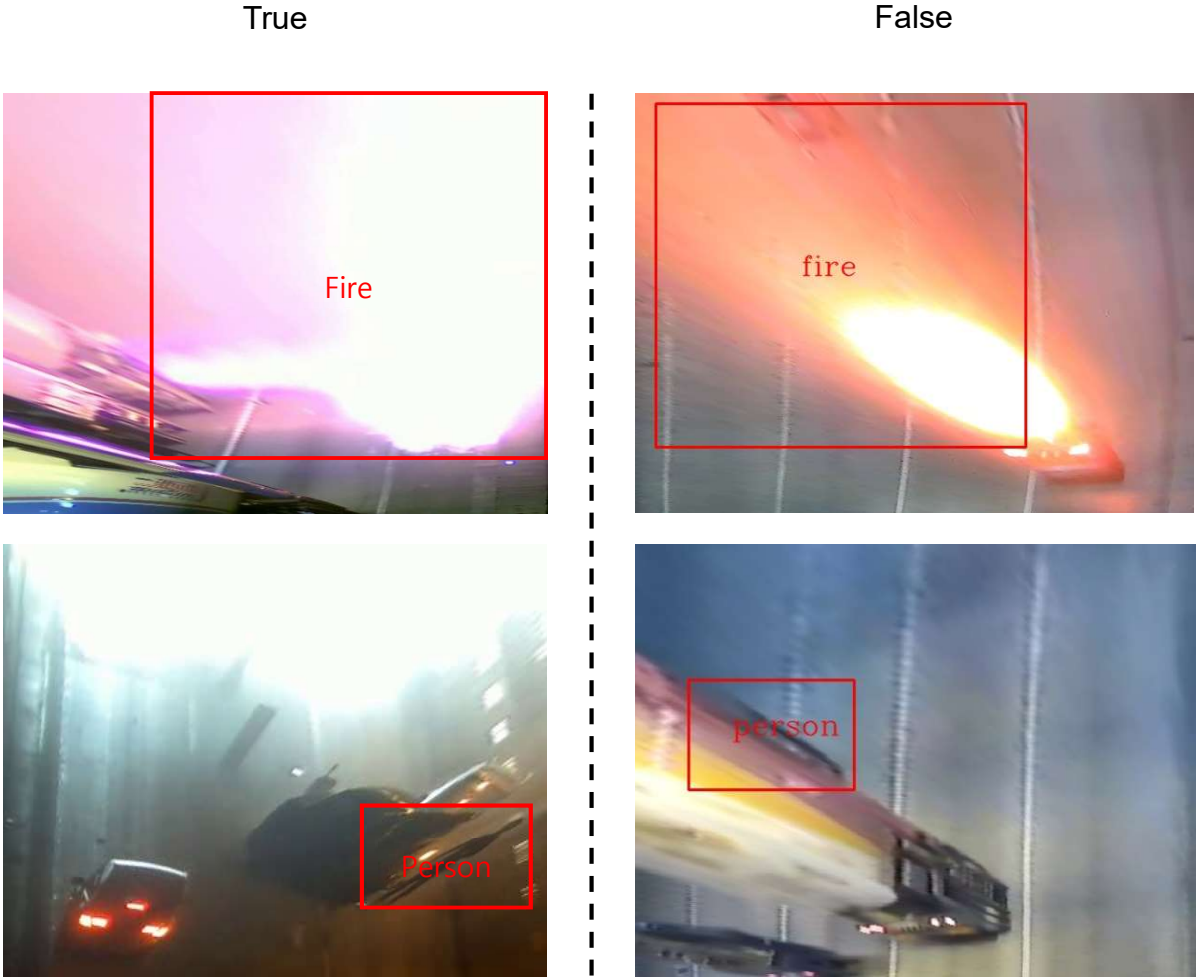

Fig. 5 Each still image sample for True and False detection on fire and person accident

Fig. 5 shows a still image sample in which a falsely detected object and a real labeled object are detected, respectively, in the fire and person object classes. The fire object is characterized by the fact that it occurs in the range of a large fire covering the whole area of the static image captured by CCTV in the case of fire in a car, and the fire object was continuously detected with time. In comparison to fire objects, person objects were much smaller and were elongated in one direction. Also, most person objects were close enough to overlap with a car, because the driver of a car gets out of the vehicle and checks the condition of the car in the event of a collision or emergency.

Recalling the characteristics of each object class, the FP samples in Fig. 5 confirm that a fire object is detected as a fire object due to the light of a car or the light reflected by a car at the entrance of the tunnel, and a person object is detected as a person in a black part of a car. Therefore, the deep-learning model trained by the labeling data was vulnerable to the detection of FPs, and a new training method of a deep-learning model that can reduce FP is needed.

## 2.2 SELF-ENHANCEMENT TRAINING METHOD with false positive data

The self-enhancement training method including FP data proposed in this paper is a method to greatly reduce the FP detection of the deep-learning model. As shown in Fig. 6, a step of collecting FP data in the training method is added to the existing deep-learning model.

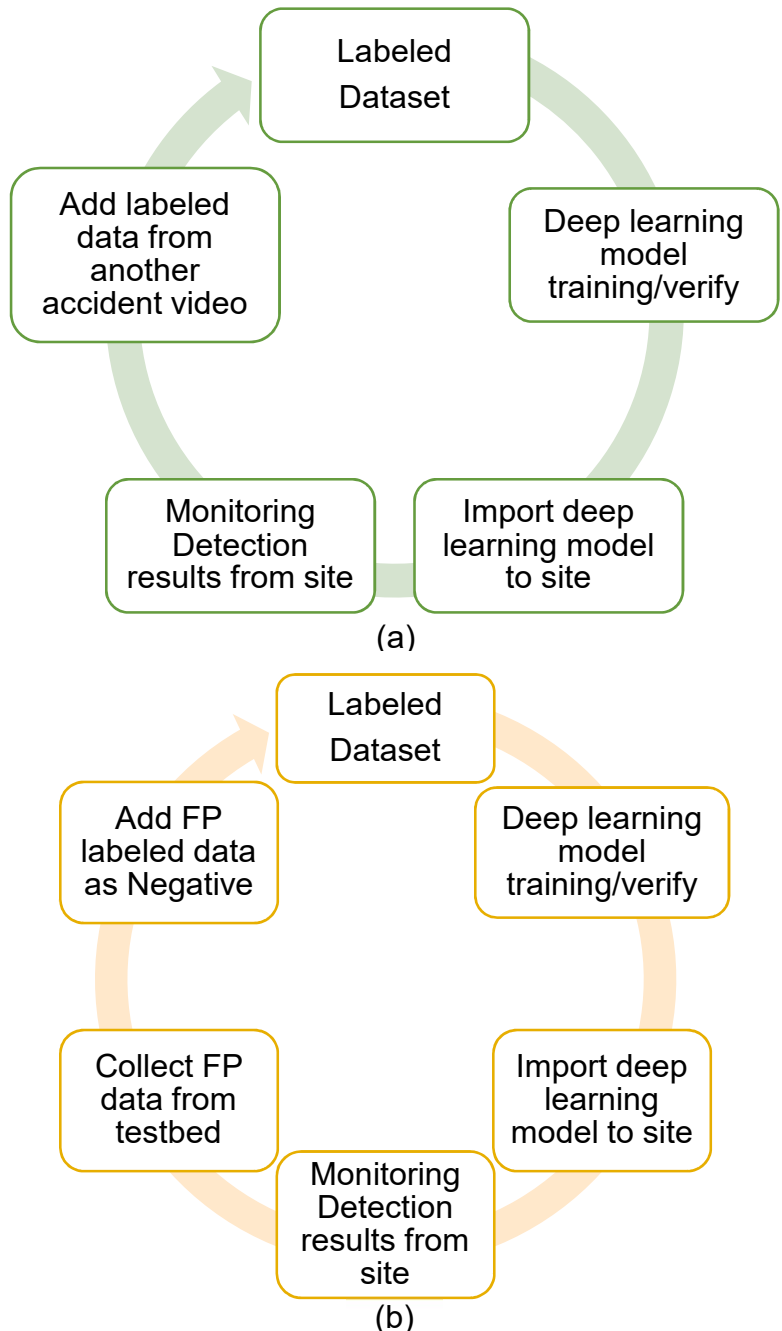

Fig. 6 The deep learning application process on TAD system for tunnel site. (a) shows traditional deep learning application process. (b) shows deep learning application process including FP data collection process

Fig. 6 (a) presents the training process of the existing deep-learning model in the TAD system. The training procedure of the existing deep-learning model first prepares the labeled dataset and then goes through the training and verification steps for the tunnel monitoring site. Then, the data of the detection results of accidents is monitored on a regular basis, and the deep-learning model applied in the field determines which object class lacks object detection ability. In the labeling step, the configuration of the still images to be added to the training data set is focused on still images of the object class that is insufficient in the monitoring step. Finally, newly labeled data is added to construct a dataset for training the deep-learning model, and this procedure can be repeated to improve the object-detection performance of the deep-learning model. However, the process of labeling data takes a considerable amount of time and effort. Also, when a still image is labeled, only the object class for which correct answers are given is trained by labeling, not the falsely detected still images. This cannot directly improve the FP results.

Fig. 6 (b) shows the self-enhancement training method including the FP data of the deep-learning object detection model proposed in this paper. Since this method requires FP data detected in the field, it was necessary to first apply the trained deep-learning model to the field with the labeling data set. Until this process, it is the same as in Fig 6 (a). However, Fig. 6 (b) shows a case in which the FP data is directly fetched in the step of monitoring the TAD system. Then the FP object is defined as a negative class, and it is added to labeled dataset to train the deep-learning model. This process is based on FP data that is not directly labeled manually. Because the deep-learning model directly trains the object class and bounding box of the FP object, in comparison with the deep-learning model trained with the existing labeling dataset, it is possible to achieve the same object-detection ability and to reduce FPs that may occur in CCTV. Thus, this method can greatly improve the field applicability and reliability of the deep-learning model.

## 3. EXPERIMENTS

In this work, we compared the FP reduction performance of the deep-learning model trained with the addition of the FP data and that of the deep learning model trained with only the labeled dataset to verify the self-enhancement training method with FP data. Accordingly, the training target labeled data was not increased, and the FP data obtained from the field was added to the training data.

For verification of the deep-learning model the detection performance for car, person, and fire object detection was assessed, and the average precision (AP), which is most widely used in object detection, was sued as an evaluation index (Zhu 2004). On the other hand, the measurement of the ability to reduce the number of FPs for person objects was evaluated by re-inferring the FP dataset used in the training of the deep-learning model, and the number of false person objects was detected. In the case of fire, the number of false alarms because of the replacement of the deep-learning model applied in the site was monitored and evaluated.

### 3.1. BIGDATA
The status of labeling data and FP data to train the deep-learning model were showed to Table 1.In the labeled dataset in Table 1, 70,914 still images were extracted from 45 accident videos, and 446,726 car objects were present in the still images. Since most of the accident data used in this work were mostly still images of driving cars, the number of car objects detected was absolutely greater than those of fire and person objects. On the other hand, there were 47,141 persons, about 1/10 of the number of car objects, even though the videos used for training were accident videos. In the case where a person appears, there is a case where a car is stopped because of a collision accident or a breakdown, and a case where the car appears for a working vehicle for tunnel work. Most of the time, one or two persons appear in a short time per still image, so it is not as easy to obtain a still image in which persons appear in comparison to the car class. In the case of fire, it is an important type of accident that appears in the news among the many types of accidents occurring in tunnels, but it is more difficult to secure videos of this type of situation because they occur less often than persons. Therefore, only two videos from 45 videos were fire accident videos, and 857 fire objects could be secured. As a result, the fire object class detected by the deep-learning model has a potential problem that does not reflect various characteristics of fires, and the light reflected at the entrance and exit of a tunnel and the warning light of a tunnel working vehicle may be falsely detected as a fire. To solve these problems, a false fire class in which non-fire light, such as the tail light of a car or the warning light of a working vehicle was defined in the labeled dataset.

Table 1. Composition of labeling data and False Positive data

| Data Category | Number of Images | Object detection category ||||| 
|---|---|---|---|---|---|---|
| | | car | person | fire | false fire | false person |
| Labeled dataset | 70,914 | 446,726 | 47,141 | 857 | 184,448 | 0 |
| FP dataset A | 2,041 | 0 | 0 | 0 | 691 | 1,357 |
| FP dataset B | 8,007 | 0 | 0 | 0 | 22 | 7,999 |

FP dataset A and FP dataset B processed the accident data detected in the tunnel for 23 days after the initial installation of the TAD system as false data. In FP dataset A, there were 691 false fire and 1357 false person from a total of 2,041 still images, and the dataset was configured to simultaneously reduce the effect of person and fire false detection. FP dataset B was composed of 7,999 false persons in a total of 8007 still images. In the case of a fire, a false fire object is simply detected as a fire due to a light object, such as a tunnel light or the warning light of a working vehicle; hence, a small number of still images can be used to reduce false fire detection. However, persons have various patterns of false person detection, such as elongated parts near cars, lanes, of parts of the tunnel wall. Therefore, FP dataset B is composed of data for the reduction of false person detection.

### 3.2. TRAINING CONDITIONS
Based on the labeling dataset and false positives dataset in Table 1, the dataset used for training of the deep-learning model is constructed as shown in Table 2.

Table 2. Data composition of each model

| Model Name | Dataset composition |
|---|---|
| model A | Labeled Dataset |
| model B | Labeled Dataset + FP dataset A |
| model C | Labeled Dataset + FP dataset A+ FP dataset B |

As seen in Table 2, model A consisted only of the labeled dataset, and it was the deep-learning model used for the initial operation of the TAD system at the tunnel site. Models B and C can be defined as training models by adding FP data detected by model A. The difference between models B and C is the scale of FP data, consisting of 2,041 still images and 10,048 still images. Model B is a model for reducing both false fire and false person detection simultaneously. Model C trains the deep-learning model by adding FP dataset A used in Model B and FP dataset B for sure reduction of FPs.

Based on these deep-learning model conditions, the training environment and variables of the deep-learning model were set as shown in Table 3.

Table 3. Deep learning model training condition

| Deep learning model | Faster R-CNN |
|---|---|
| GPU | NVIDIA GTX 1070 |
| OS | Linuxmint 18.3 |
| Epoch | 10 |
| Learning rate | 0.001 |
| Convolutional layer | VGGnet 16 layer |

Table 3 shows the faster regional convolutional neural network (R-CNN), the deep-learning model used in this paper. Faster R-CNN is a much faster and more accurate deep-learning object-detection network than existing deep-learning object-recognition networks at the time of raster R-CNN's announcement (Ren 2015). As of 2019, there is a better deep-learning object detection network than faster R-CNN, but faster R-CNN is widely used in various research fields because it is a deep-learning network that is a reference point in deep-learning object detection (Zhao 2019). Therefore, this work also uses faster R-CNN. The GPU uses the NVIDIA GTX 1070 and can learn at an average rate of 0.215 seconds per still image.

In the training variable, the epoch is set to 10, and the learning rate is 0.001. Finally, the convolutional layer, which greatly affects learning and reasoning speed and accuracy in faster R-CNN, uses a 16-layer Visual Geometry Group (VGG) network. The VGG network is accurate enough to be ranked second in the ILSVRC 2014 image classification competition (Simonyan 2015). The structure of the network is 3 x 3 for the convolutional filter and 2 x 2 for the pooling layer, making it easy to understand and apply. Therefore, the VGG network is used for various studies, and in this work, it is also used as the convolutional layer of faster R-CNN.

## 3.3. EXPERIMENTAL RESULTS

First, the re-inference object detection performance for each deep-learning model is shown in Fig. 7.As seen in Fig. 7, the average precision (AP) for car objects are excellent values of 0.85 to 0.88 for all of the three deep learning models, and since a large number of car objects have been trained, it is possible to detect a car object without any problem when it actually enters the tunnel site. The object-detection performance for the person class was improved by training with the addition of false persons. In the case of model A, the AP value of 0.72 was detected, but the value of model B was 0.74, and the value of model C was 0.77. This tendency shows that the false-detection reduction the deep learning model can improve the object detection ability. The fire object class had a high AP value of 0.91 for all three deep-learning models because the number of fire objects was as low as 857, which is easy to learn about the fire object class. Therefore, for the fire object class, the detection trend of FPs can be assessed by applying the model B trained with the addition of false fires to the site where the TAD system is installed.

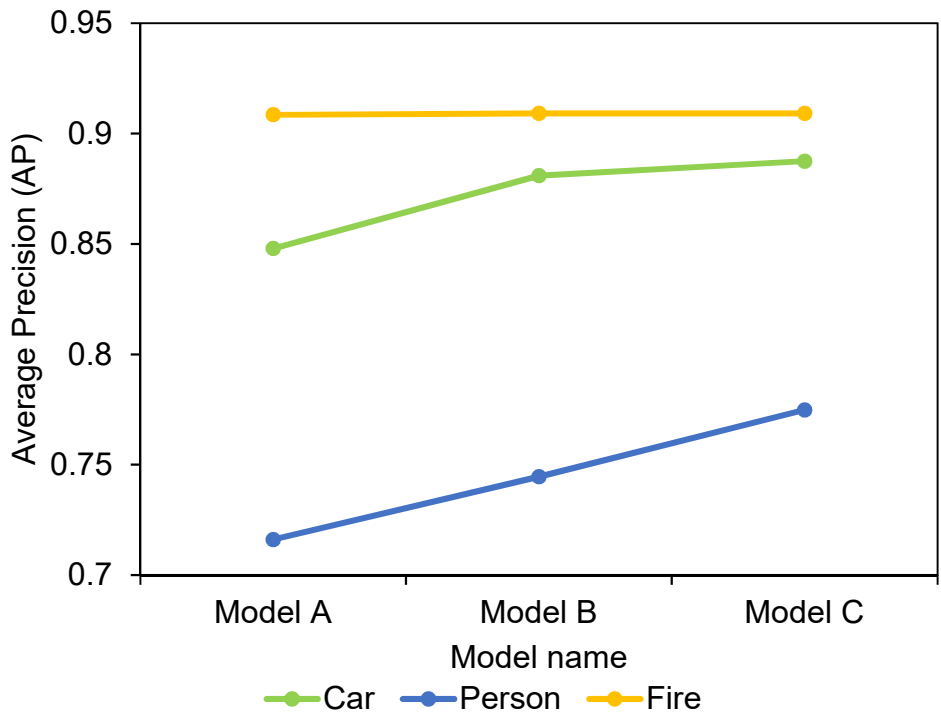

Fig. 7 The deep learning model object detection performance for 3 object class

Table 4. The number of false positives detected by inferring trained False Positive data in each model

| Target Model | Rate of detected person as FP in FP dataset ||
|---|---|---|
| | FP dataset A | FP dataset A+B |
| Model B | 16.12% | 32.54% |
| Model C | 1.67% | 4.64% |

Table 4 shows the percentage of false persons detected by re-inferring the FP data of person sets for models B and C trained with the FP dataset. First, when model B was re-inferred from the FP dataset A, the rate of detection of false person was 16.12% for all still images, and for FP datasets A and B the rate was 32.54%. On the other hand, model C was detected as 1.67% for FP dataset A, and it was 4.64% for FP datasets A and B; compared with model B, the detection of false persons was reduced. Therefore, to determine the effect of the false person reduction, about 9000 false persons should be added to the training dataset when training the deep-learning model. False fire detection was performed by monitoring model A for the first 56 days after addition of model A in the TAD system installed at the tunnel site and then updating the trained model B by adding the false fire dataset. In the TAD system installed at the tunnel site, the detection status of false fire alarms of 9 channels of CCTV is shown in Fig. 8.

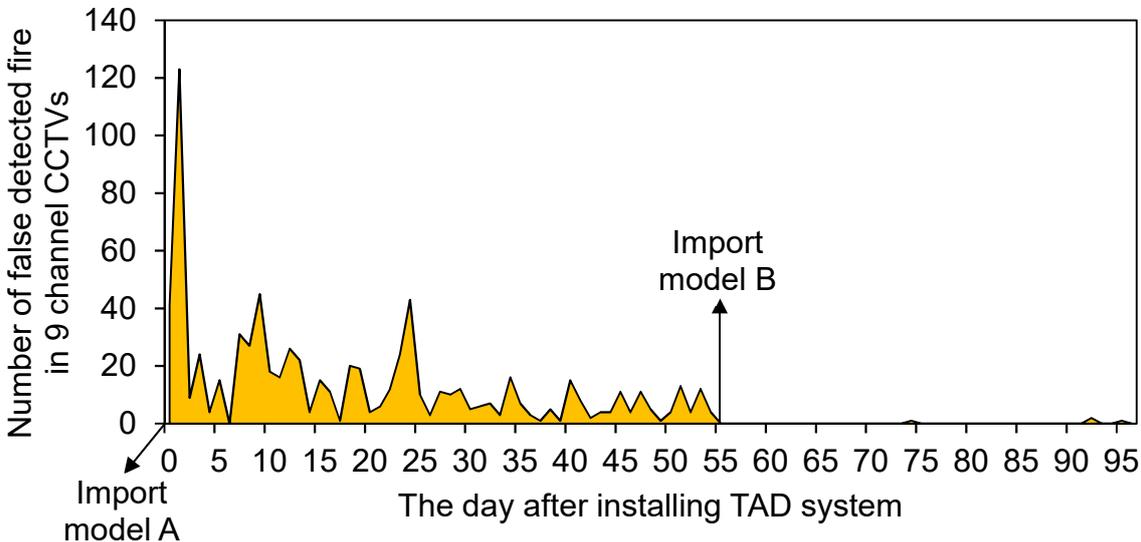

Fig. 8 The number of false detected fire on TAD system per day. The model B was updated to 56 days after installing TAD system

Initially, the TAD system was installed in the tunnel site as shown in Fig. 8, and model A was applied to the TAD system and monitored for 56 days. In this case, false fires consistently occurred with a maximum of about 120, usually 0 to 40. Then, after application of model B to the TAD system, monitoring for 40 days showed false fires in the range of 0 to 2, which was a dramatic reduction in comparison to model A. The result is that the number of the fire object class was not sufficient to reduce diversity. The patterns of false fires detected in the field was also affected by the light from the tunnel entrance and exit, so the simple types were detected. Since these types of false fire objects have a larger size in comparison to persons and cars, the trained deep-learning model including the false fire object class can relatively easily reduce false fire detection in tunnel sites.

## 4. CONCLUSION

In this paper, the self-enhancement training method including FP data was introduced and compared with a deep-learning model that is trained only with the existing labeled dataset. The object detection ability using the labeled dataset, the false detection rate of person with re-inference, and the number of false fires detected by monitoring in the tunnel site were assessed. Analysis of each task revealed that the object detection ability is maintained or improved when FP detection data is included in training. In addition, the rate of false person detection decreases as the number of false persons included in the training dataset increases. In addition, in the TAD system installed at the tunnel site, when we compared the deep-learning model installed and the deep learning model trained with false fire data, the proposed method achieved better results. Based on these results, the following conclusions were obtained.

First, this work demonstrated effect of the self-enhancement training method including FP data, but it did not directly compare the effects of the existing deep-learning model training procedure. Therefore, there is a need for research that can directly compare two methods of deep-learning training.

2. In the tunnel CCTV labeled dataset to be used for the training of the deep-learning model, the tendency of the FP detection changes according to the size and characteristics of each object class. The fire class was very easy to train because the scale of the fire data was very small compared to car and person data, and it showed a relatively large object size and light shape. As a result, false fires are also expressed in the form of large object size and light; thus, the number of FPs for the fire object class is easy to reduce. On the other hand, in the case of the person class, the amount of data is much greater than for the fire class, but the size of the objects is much smaller than that of fire objects, and training of deep learning is not easy. Since many types of false persons were detected on-site in the TAD installed in the field, the false person class must train at least 9000 false person objects in order to achieve false person detection reduction.

3. The deep-learning self-enhancement training method including FP data can improve the object detection ability with reducing the FPs. Therefore, it is possible to increase the reliability of the deep-learning model when applying it in the field, and it can be useful not only in the TAD system considered in this paper but also in other deep-learning based applications.